# Real-Time Action Detection in Video Surveillance using Sub-Action Descriptor with Multi-CNN

Cheng-Bin Jin[*], Shengzhe Li[†], and Hakil Kim[*]

[*]Inha University, Incheon, Korea
[†]Visionin Inc., Incheon, Korea

## Abstract

When we say a person is texting, can you tell the person is *walking* or *sitting*? Emphatically, no. In order to solve this incomplete representation problem, this paper presents a sub-action descriptor for detailed action detection. The sub-action descriptor consists of three levels: the posture, the locomotion, and the gesture level. The three levels give three sub-action categories for one action to address the representation problem. The proposed action detection model simultaneously localizes and recognizes the actions of multiple individuals in video surveillance using appearance-based temporal features with multi-CNN. The proposed approach achieved a mean average precision (mAP) of 76.6% at the frame-based and 83.5% at the video-based measurement on the new large-scale ICVL video surveillance dataset that the authors introduce and make available to the community with this paper. Extensive experiments on the benchmark KTH dataset demonstrate that the proposed approach achieved better performance, which in turn boosts the action recognition performance over the state-of-the-art. The action detection model can run at around 25 fps on the ICVL and more than 80 fps on the KTH dataset, which is suitable for real-time surveillance applications.

Keywords: sub-action descriptor, action detection, video surveillance, convolutional neural network, multi-CNN

## 1. Introduction

The goal of this paper is to enable automatic recognition of actions in surveillance systems to help human for alerting, retrieval, and summarization of the data [1], [2], [3]. Vision-based action recognition—the recognition of semantic spatial–temporal visual patterns such as *walking*, *running*, *texting*, and *smoking*, etc.—is a core computer vision problem in video surveillance [4]. Much of the progress in surveillance has been possible owing to the availability of public datasets, such as the KTH [5], Weizmann [6], VIRAT [7], and TRECVID [8] datasets. However, current state-of-the-art surveillance systems have been saturated by these existing datasets, where actions are in constrained scenes and some unscripted surveillance footage tends to be repetitive, often dominated by scenes of people walking. There is a need for a new video surveillance dataset to stimulate progress. In this paper, the ICVL dataset[1] is introduced, which is a new large-scale video surveillance dataset designed to assess the performance of recognizing an action and localizing the corresponding space–time volume from a long continuous video. The ICVL dataset has an immediate and far-reaching impact for many research areas in video surveillance, including human detection and tracking and action detection.

---

[1] Available from: http://vision.inha.ac.kr/

For the existing representation problem of an action, this paper proposes a sub-action descriptor (Figure 1), which delivers complete information about a human action. For instance, conventional methods give the action representation of *texting* for one person who is *texting while sitting* and the same action descriptor for another person who is *texting while walking*. The action information for those two persons should be totally different. The first difference is posture: one person is *sitting*, and the other is *standing*. The second difference is locomotion: one person is *stationary*, and the other is *walking*. The proposed sub-action descriptor consists of three levels: posture, locomotion, and gesture, and the three levels provide three sub-action descriptors for one action to address the above problem. Each level of sub-action descriptor is one convolutional neural network (CNN)-based classifier. Each CNN classifier captures different appearance-based temporal features to represent a human sub-action.

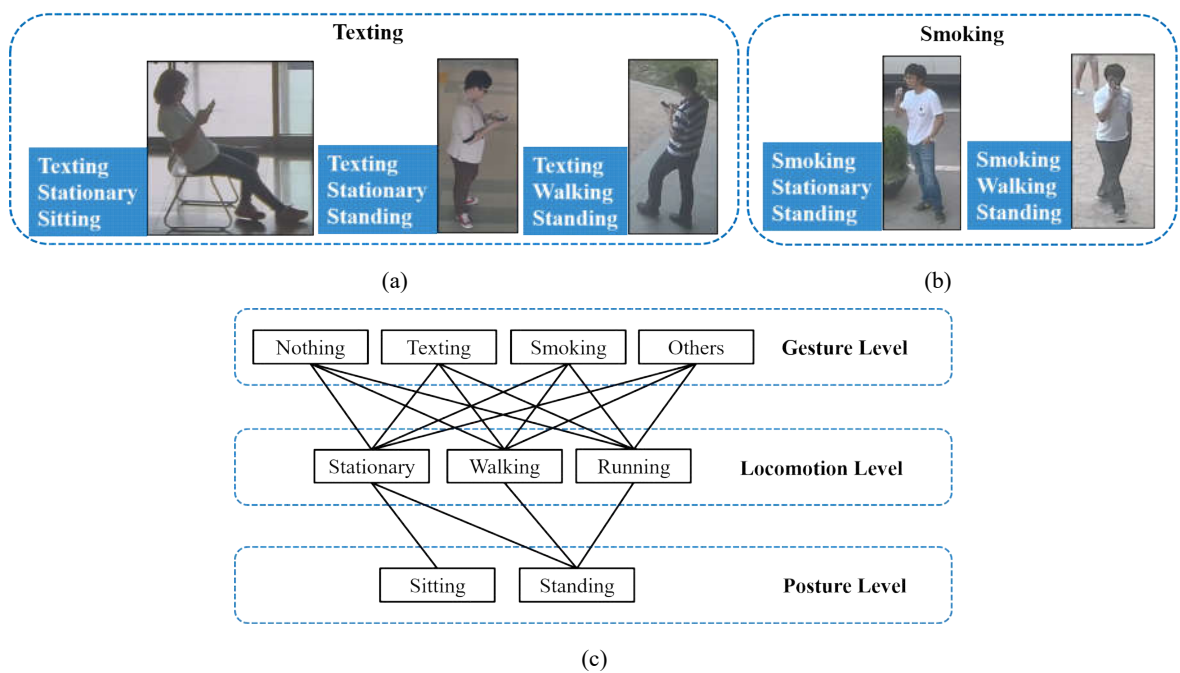

Figure 1: Conventional representation problems and sub-action descriptor. (a) Conventional representation problem: *texting*, (b) conventional representation problem: *smoking*, and (c) structure of the sub-action descriptor. A sub-action descriptor includes three levels: the posture level, the locomotion level, and the gesture level. Each level has one CNN; therefore, for one action by an individual, three CNNs work simultaneously.

Much of the existing work [9], [10] is focused on video-based action recognition ("*Is there an action in the video?*") that tries to classify the video as a whole via globally pooled features. Global-based feature pooling works well; however, this method does not consider the different actions of multiple individuals that are present at the same time. For instance, one person in the video might be *texting* and, beside him, another person is *smoking*. In our work, the problem of action detection in video surveillance is addressed: "*Is there an action, and where is it in the video spatially and temporally?*" This does far more than most of the works so far, which aims to classify a pre-clipped video segment of a single action event. The rationale behind the action detection strategy is partly inspired by the technique used in a recent paper [11], which starts by localizing action interest regions and classifying them, which improves the representational power and classification accuracy.

This paper aims to develop a real-time action detection algorithm with high performance based on the CNN. This is challenging, since tracking by detection and action recognition are computationally expensive and cannot be estimated together in real-time. There are many works to estimate human pose [12], [13], [14] and analyze motion information [15] in real-time. B. Zhang, et al. [16] proposed a real-time CNN based action recognition method. However, to the best of our knowledge, none of the work can spatially and temporally detect actions for real-time process.

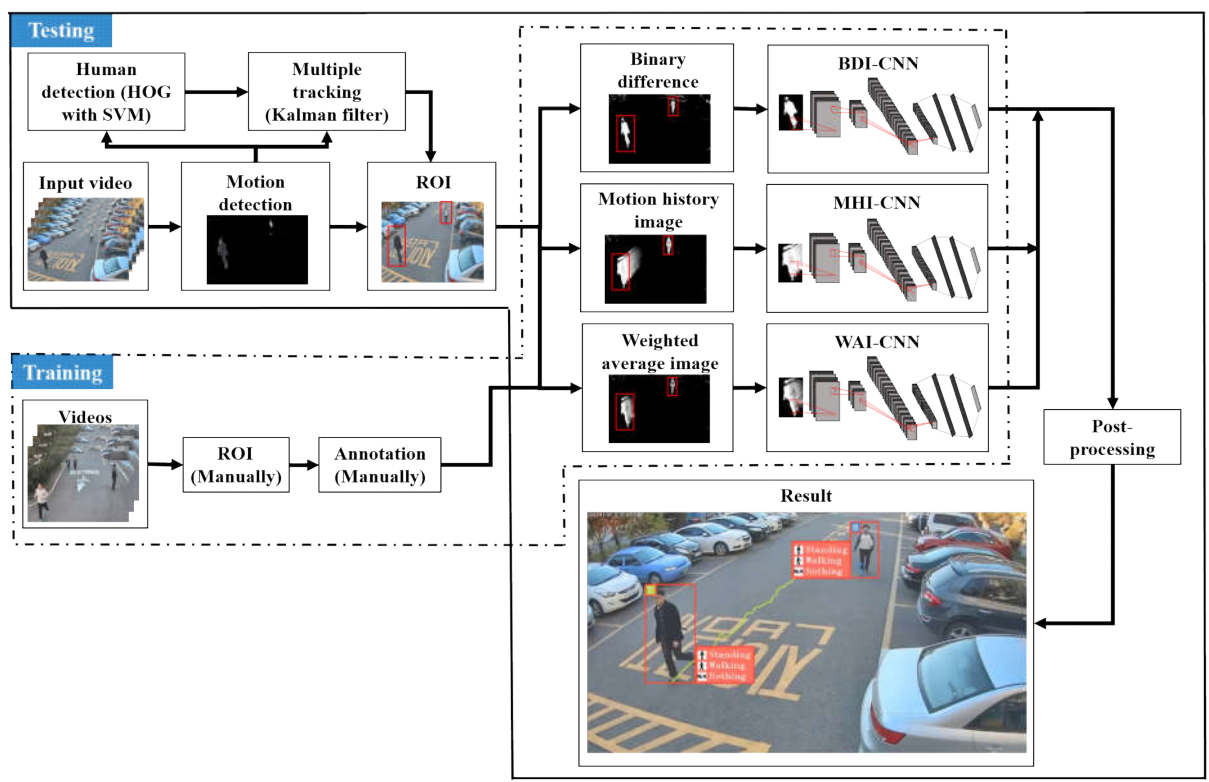

Figure 2: Overall scheme of the proposed real-time action detection model. Through a motion-detection, human-detection, and multiple-tracking algorithm, appearance-based temporal features of the regions of interest (ROIs) are fed into three CNNs, which make predictions using shape, motion history, and their combined cues. Each action of an individual has three sub-action categories under the sub-action descriptor, which delivers a complete set of information about a human action.

The goal of this work was to build a novel model that can simultaneously localize and recognize multiple actions of individuals in video surveillance. Figure 2 shows the overall scheme of the proposed real-time action detection model. In the training phase, the ROI and the sub-action annotations are first determined manually in each frame of the training videos, and three appearance-based temporal features—binary difference image (BDI), motion history image (MHI), and weighted average image (WAI)—are computed from the ROI. Every single level of the sub-action descriptor has one CNN classifier, and this paper denotes the classifiers as BDI-CNN, MHI-CNN, and WAI-CNN. These classifiers are learned via three appearance-based temporal features. In the multi-CNN model, each prediction of a CNN is equal to one sub-action. In the testing phase, a motion saliency region is generated using a Gaussian mixture model (GMM) to eliminate regions that are not likely to contain the action. This leads to a big reduction in the number of regions processed. The conventional sliding window–based scheme is used on the motion saliency region as a mask. In the sliding window, a human-detection histogram of oriented gradient (HOG) descriptor [17] with a latent support vector machine (SVM) [18] is used to detect humans in an initial action in the ROIs. Then, the regions undergo Kalman filtering–based refinement of the positions. Given the refined action in the regions of interest, shape, motion history, and their combined cues are used with the aid of the CNNs to predict three sub-action categories. Finally, the post-processing stage checks for conflicts in the structure of the sub-action descriptor and applies temporal smoothing according to the previous action history of each individual to reduce noise.

Experimental results are shown for the task of action detection with the ICVL dataset and KTH dataset. An ablation study is presented and shows the effect of each component when considered separately. The results indicate that shape and motion history information are complementary, and that using both leads to a significant improvement in performance for recognizing subtle actions. The proposed approach achieves a mean average precision (mAP) of 76.6% at the frame-based measurement and 83.5% at the video-based measurement with the ICVL dataset. Moreover, the results on the KTH demonstrate that the proposed method significantly outperforms state-of-the-art action recognition methods. This paper builds on an earlier publication [19], which unifies the notation, explains the approach in more detail, and includes considerably more thorough experimental validation. The major contributions of this paper can be summarized as follows:

- The sub-action descriptor is described for the action detection model. In this descriptor, there are three levels; the levels are combined to represent many different types of actions with a large degree of freedom. The use of divided levels in the sub-action descriptor delivers a complete set of information about human actions and is based on the experimental results, which significantly eliminates misclassifications.
- A real-time action detection model is developed on the basis of appearance-based temporal features with a multi-CNN classifier. Much of the work in human-activity analysis focuses on video-based action classification. However, a model for action detection that simultaneously localizes and recognizes multiple actions of individuals with low computational cost and high accuracy is provided.
- A new public surveillance video dataset is introduced. A collective effort was made to obtain natural examples from a variety of sites under different weather conditions and levels of illumination. Detailed annotations are available, which include human bounding box tracks and sub-action labels, which provide quantitative evaluation for surveillance research. This is the only dataset suitable for action detection in surveillance, unlike datasets for the task of action classification.

The rest of this paper is organized as follows. More background information about activity analysis is provided in Section 2. The details of the proposed approach are described in Section 3. An evaluation and a discussion of the proposed method performed on the ICVL and KTH dataset are given in Section 4. Finally, conclusions, including potential improvements, are given in Section 5.

## 2. Related Works

There has been a fair amount of research on activity analysis, and recent surveys can be found [20], [21]. Estimating a human pose using a predefined model (e.g., pictorial structures) in each frame is the most common technique for generic human model recovery. The model is driven by an attempt to minimize the cost function between the collection of parts arranged in a deformable configuration and human contours [22], [23]. Because of the truncation and occlusion of body parts, pose estimation from a 2D image is a complicated process. Recent work has exploited improvements in depth imaging and 3D input data. Shotton et al. [24], [12] estimated parts of the human body and the 3D locations of each skeletal joint directly from a single depth image using a 3D sensor. However, a human action goes further than a human pose, and after human pose information is obtained, specific classifiers or decision rules are needed to recognize actions.

Many of the approaches introduce an appearance-based method where an action comprises a sequence of human silhouettes or shapes. In contrast to human model recovery, this method uses appearance-based temporal representations of static cues in a multitude of frames, where an action is described by a sequence of two-dimensional shapes. Motion energy image (MEI) and motion history image (MHI) [25], [26] are the most pervasive appearance-based temporal features. The advantages of the methods are that they are simple, fast, and work very well in controlled environments, e.g., the background of the surveillance video (from a top-view camera) is always the ground. The fatal flaw in MHI is that it cannot capture interior motions; it can only capture human shapes [19]. However, the effect of shape and motion history cues with CNN for action recognition has not been investigated carefully. In our work, a novel method for encoding these temporal features is proposed, and a study of how different appearance-based temporal features affect performance is provided. Other appearance-based temporal methods are the active shape model, the learned dynamic prior model, and the motion prior model. In addition, motion is consistent and easily characterized by a definite space–time trajectory in some feature spaces. Based on visual tracking, some approaches use motion trajectories (e.g., generic and parametric optical flow) of predefined human regions or body interest points to recognize actions [27], [28].

Local spatial–temporal feature-based methods have been the most popular over the past few years. The methods compute multiple descriptors, including appearance-based (e.g., HOG [17], Cuboids [29] or SIFT [30]) and motion-based (e.g., optical flow [31], HoF [32], MBH [33]) features on a spatial–temporal interest point trajectory, which they encode by using a bag of features or Fischer vector encoding [34] and by training SVM classifiers. Laptev [35] proposed space-time interest point (STIP) by extending the 2D Harris corner to a 3D spatial–temporal domain. Kim et al. [36] introduced a multi-way feature pooling approach that uses unsupervised clustering of segment-level HoG3D [37] features. Li et al. [38] extracted spatial–temporal features that are a subset of improved dense trajectory (IDT) features [10], [33], namely, HoF, $MBH_x$, and $MBH_y$, by removing camera motion to recognize egocentric actions. The local spatial–temporal feature-based method was shown to be efficient with challenging scenes and achieved state-of-the-art performance with several instances of benchmark action recognition. However, the existing methods are quite computationally expensive.

Some alternative methods for action recognition have been proposed [11], [39], [40], [41]. Vahdat et al. [39] developed a temporal model consisting of key poses for recognizing higher-level activities. Lan et al. [40] introduced a structure for a latent variable framework that encodes contextual information. Jiang et al. [11] proposed a unified tree-based framework for action localization and recognition based on a HoF descriptor and a defined initial action segmentation mask. Lan et al. [41] introduced a multi-skip feature-stacking method for enhancing the learnability of action representations. In addition, hidden Markov models (HMMs), dynamic Bayesian networks (DBNs), and dynamic time warping (DTP) are well-studied methods for speed variation in actions. However, actions cannot be reliably estimated in real-world environments using these methods.

Computing handcrafted features from raw video frames and learning classifiers on the basis of the obtained features are a basic two-step approach used in most of the methods. In real-world applications, the design of the feature and the choice of the feature are the most difficult and highly problem-dependent issues. Especially for human action recognition, different action categories may look dramatically different according to their appearances and motion patterns. Based on impressive results from deep architectures, attempts have been made to train deep convolutional neural networks for the task of action classification [42],[43]. Ji et al. [44] built a 3D CNN model that extracts appearance and motion features from both spatial and temporal dimensions in multiple adjacent frames. Using two-stream deep convolutional neural networks with optical flow, Simonyan and Zisserman [45] achieved a result that is comparable to IDT [10]. Karpathy et al. [46] trained a deep convolutional neural network using 1 million videos for action classification. Gkioxari and Malik [47] built action detection models that start by selecting candidate regions using CNNs and then classify them using SVM. The two-stage structure in our proposed approach for action detection is similar to their work; however, the crucial difference is that their network focuses on one actor and made incorrect predictions for multiple actors based on their optimization problem. Moreover, appearance-based temporal-feature integration is quite different, and our proposed approach is able to detect the actions of multiple actors.

## 3. Proposed Model for Human Action Detection

The main objective of the proposed approach is to detect the actions of multiple individuals for real-time surveillance applications. Figure 2 outlines the proposed approach. The human action regions are detected by a frame-based human detector and a Kalman tracking algorithm. The action classifier is composed of three CNNs that operate on the shape, motion history and their combined cues. According to the sub-action descriptor, the classifier predicts the regions to produce three outputs for each action. The outputs of the classifiers go through a post-processing step to render the final decisions.

### 3.1 Sub-Action Descriptor

The problem of representing an action is not well-defined as a measurement problem of geometry (e.g., measurement of an image or camera motion). Intra-class variation in the action category is ambiguous, as shown in Figure 1(a) and (b). Although the actions of the three persons are *texting* in Figure 1(a), can you tell if what they are doing is exactly the same? The first person is *texting while sitting*, the second person is *texting while standing* and is *stationary*, and the third person is *texting while standing* and *walking*. For the above three persons, giving the same action representation (*texting*) is often confusing—all of them have different postures and locomotion states for the same action. These are the same problems for the action of *smoking* in Figure 1(c).

To deliver complete information about human actions, and to clarify action information, the proposed approach in this paper models an action with a sub-action descriptor. A depiction of the sub-action descriptor is shown in Figure 1(c). The descriptor includes three levels: the posture level, the locomotion level, and the gesture level. The posture level comprises the following two sub-actions: *sitting* and *standing*. The locomotion level comprises *stationary*, *walking*, and *running*. The gesture level comprises *nothing*, *texting*, *smoking*, and *others* (e.g., *phoning*, *pointing*, or *stretching*, which are not considered).

The connecting line between two sub-actions at different levels indicates that the two sub-actions are independent of each other. No connection indicates an incompatible relation where the two sub-actions cannot happen together. Each level has one CNN; therefore, for one action of an individual, three CNNs work simultaneously. The first network, BDI-CNN, operates on a static cue and captures the shape of the actor. The second network, MHI-CNN, operates on a motion cue and captures the history of the motion of the actor. The third network, WAI-CNN, operates on a combination of static and motion cues and captures the patterns of a subtle action by the subject. The designed descriptor of actions can transform the difficult problem of action recognition into many easier problems of multi-level sub-action recognition. It is inspired by the very large number of actions that can be built by very few independent sub-actions. In this descriptor, three levels combine to represent many different types of actions with a large degree of freedom.

### 3.2 Tracking by Detection

The main goal of this paper is real-time action detection in surveillance video. For the human detection and tracking algorithm, we adopt existing methods to provide a stable human action region for subsequent action recognition. A processing time of 20-30 *ms* for each frame, a stable bounding box for the human action region, and a low false detection rate are the important factors for human detection and tracking.

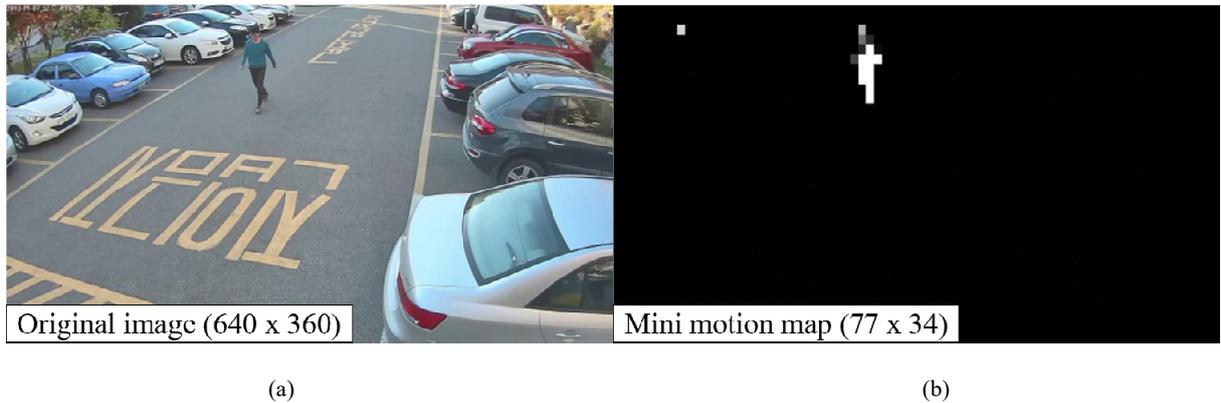

(a)          (b)

Figure 3: Mini motion map for reducing the unnecessary computation in HOG-based human detection. (a) Original image with a size of 640 × 360, and (b) mini motion map with a size of 77 × 34, which was calculated from GMM-based motion detection

Computational efficiency of the surveillance application is critical. Sliding window is the bottleneck in the processing time of the object detection because many windows, in general, contain no object. To this end, motion detection is performed before object detection to discard regions that are void of motion. A mini motion map is generated [48] by using a Gaussian mixture model–based motion detection algorithm [49]. The mini motion map is rescaled to the same number of width × height sliding windows, such that minimum computation is required when determining whether a sliding window is a foreground or a background. The size of the mini motion map is computed with the following equation:

$$\textbf{size}_{\text{mni-map}} = \frac{\textbf{size}_{\text{original}} - \textbf{size}_{\text{detection}}}{\textbf{stride}} \quad . \tag{1}$$

The default value of **size**$_{detection}$ is (64, 128) and that of **stride** is (8, 8) in HOG [17]. Figure 3 shows the mini motion map. For instance, if the size of the original image is 640 × 360, then the size of the mini motion map is 77 × 34.

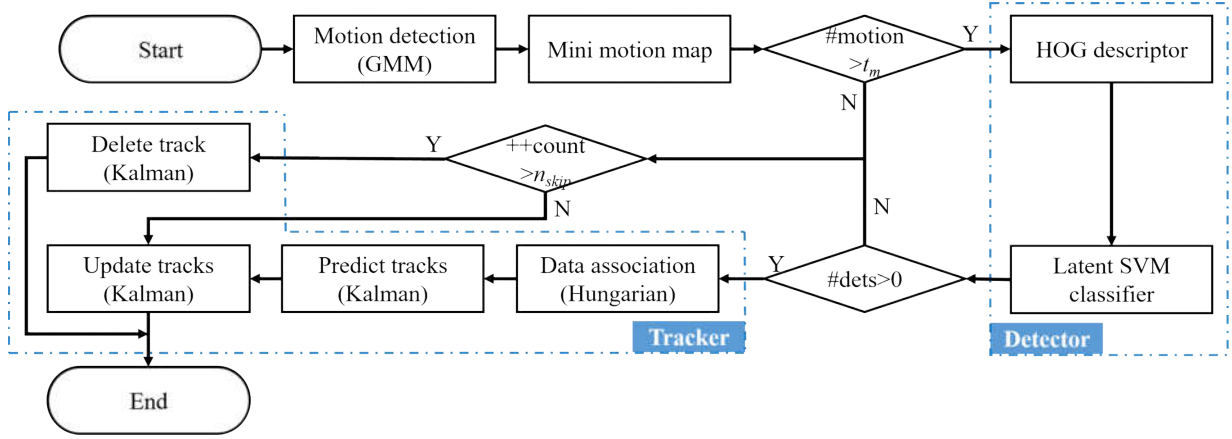

Figure 4: Procedure for multiple detections and tracks.

For the classification, latent SVM [18] with a fixed $C = 100$, as recommended elsewhere [10], is used to classify the HOG to detect humans. The next step is to associate the resulting detections with the corresponding tracks [50]. Humans appear and disappear at random times; thus, three types of cases exist in the data association problem: 1) add a new track, 2) update an existing track, and 3) delete a track [48]. The procedure for handling multiple detections and tracks is shown in Figure 4. When a new track is added, it starts to count the number of frames that the track has updated without detection. In this way, even when a motion is not detected in some frames, the track still updates according to the previous prediction. If the number is larger than the threshold $n_{skip}$, the track is considered to have disappeared and is deleted.

### 3.3 Appearance-Based Temporal Features

Appearance-based temporal features extract static information from a multitude of frames, where an action is described by a sequence of two-dimensional shapes. The features are very simple and fast, and they work very well in controlled environments, such as in surveillance systems where the cameras are installed on rooftops or high poles; therefore, the view angles of the cameras are toward dominant ground planes.

For the present discussion, a video $F$ is just a real function of three variables:

$$F = f(x, y, t) \qquad . \qquad (2)$$

Here, the coordinate $(x, y, t) \in R^3$ is the Cartesian coordinate of the video space. In a sub-action descriptor, each level has one independent CNN that obtains different appearance-based temporal features. The BDI feature accurately captures the static shape cue of the actor in 2D frames, denoted as $b(x, y, t)$, and is given by Eq. 3:

$$b(x,y,t) = \begin{cases} 255, & \text{if } f(x,y,t) - f(x,y,t_0) > \xi_{thr} \\ 0, & \text{otherwise} \end{cases}, \quad (3)$$

where the values in the BDI are set to 255 if the difference between the current frame $f(x, y, t)$ and the background frame $f(x, y, t_0)$ of the input video is bigger than a threshold $\xi_{thr}$, and $x$ and $y$ are indexes in the image domain. BDI is a binary image that indicates the silhouette of the posture. Examples are given in Figure 5.

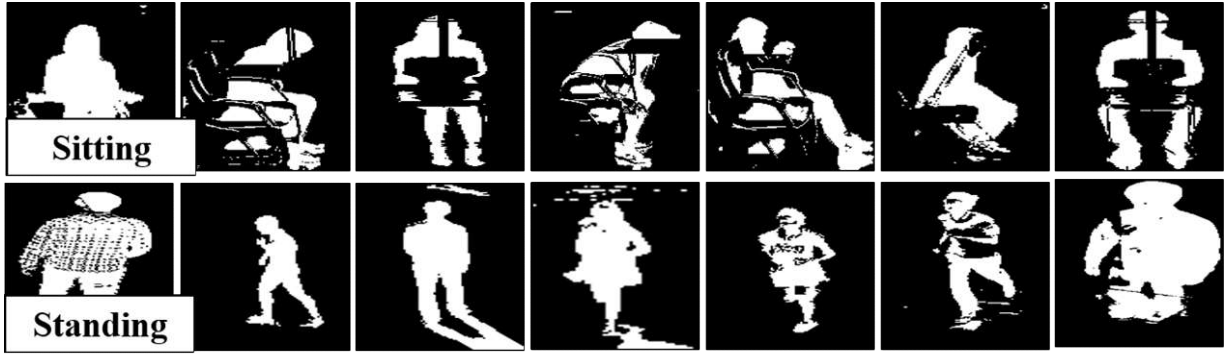

Figure 5: Examples of BDI for different sub-actions. BDIs are utilized for the posture level of the sub-action descriptor, which comprises *sitting* and *standing*. BDI captures the static shape cue of the actor.

In a motion history image, pixel intensity is a function of the temporal history of motion at that point. MHI captures the motion history patterns of the actor, denoted as $h(x, y, t)$, and is defined using a simple replacement and decay operator in Eqs. 4-6 [25]

$$d(x,y,t) = \begin{cases} 255, & \text{if } f(x,y,t) - f(x,y,t-1) > \xi_{thr} \\ 0, & \text{otherwise} \end{cases} \quad (4)$$

$$h(x,y,t) = \begin{cases} \tau_{max}, & \text{if } d(x,y,t) = 255 \\ max(0, h(x,y,t-1) - \Delta\tau) & \text{otherwise} \end{cases} \quad (5)$$

$$\Delta\tau = \frac{\tau_{max} - \tau_{min}}{n}. \quad (6)$$

MHI is used for the locomotion level, which comprises *stationary*, *walking*, and *running*. It is generated from the difference between the current frame $f(x, y, t)$ and the previous frame $f(x, y, t-1)$ in Eq. 4. For each frame, the MHI at time $t$ is calculated from the result of the previous MHI. Therefore, this temporal feature does not need to be calculated again for the whole set of frames. MHI is a vector image of motion, where more recently moving regions are brighter (see Figure 6). In Eq. 6, $n$ is the number of frames to be considered as the action history capacity. The hyper-parameter $n$ is critical in defining the temporal range of an action. An MHI with a large $n$ covers a long range of action history; however, it is insensitive to current actions. Similarly, MHI with a small $n$ puts the focus on the recent actions and ignores past actions. Hence, choosing a good $n$ can be fairly difficult.

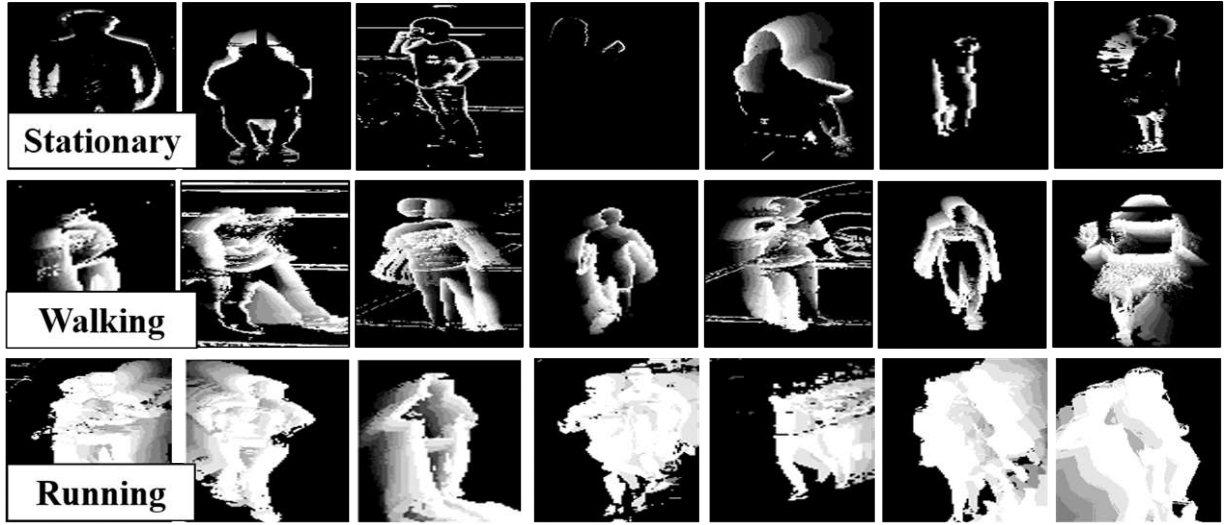

Figure 6: Examples of MHI for different sub-actions. MHIs are used for the locomotion level of the sub-action descriptor, which comprises *stationary*, *walking*, and *running*. MHI captures the motion history cue of the actor, where more recently moving pixel regions are brighter. The human eye can easily distinguish *stationary*, *walking*, and *running* from MHIs.

Weighted average images (WAIs) are applied at the gesture level of the sub-action descriptor, which comprises *nothing*, *texting*, *smoking*, and *others*. For recognizing subtle actions (e.g., *texting* and *smoking*), the easiest way would be to use the shape or motion history of the actor. The problems with this method are that it cannot capture detailed information about the subtle actions or that it is sensitive to context movement, such as camera vibration. The combined cues of shape and motion history together obtain a spatial–temporal feature for subtle actions. WAI is denoted as $s(x, y, t)$. It is constructed as a linear combination of BDI and MHI, given by Eq. 7:

$$s(x,y,t) = w_1 \cdot b(x,y,t) + w_2 \cdot h(x,y,t), \quad \text{s.t.} \ w_1 + w_2 = 1 \qquad . \tag{7}$$

As actions become more complicated, WAI is still not lost completely. $\mathbf{w} = \{w_1, w_2\}^T$ is another hyper-parameter. Figure 7 shows some examples of WAI for different sub-actions.

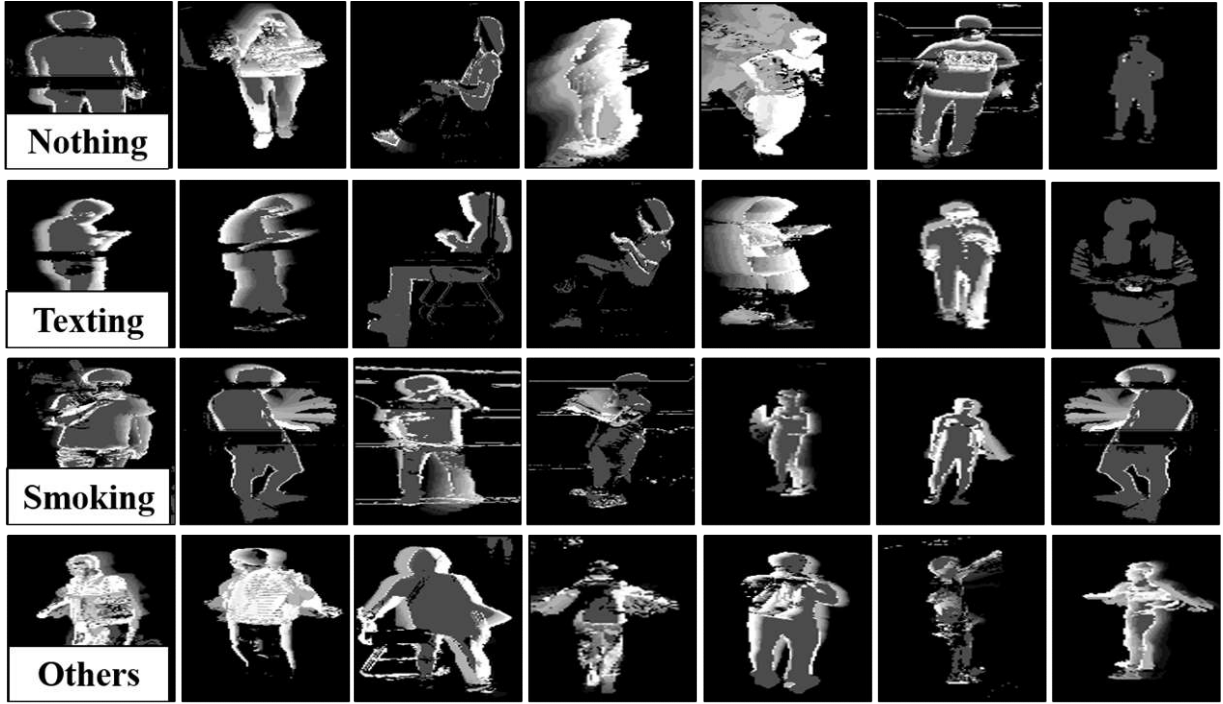

Figure 7: Examples of WAI for different sub-actions. WAIs were applied at the gesture level of the sub-action descriptor, which comprises *nothing*, *texting*, *smoking*, and *others*. WAI obtained the combined cues of shape and motion history. *Texting* (frequently moving fingers) and *smoking* (repeated hand-to-mouth motion) were captured in WAIs.

The design of these appearance-based temporal features is strongly motivated by their computation: fast, memory efficient, and with no preprocessing step. Given a larger computational budget, one could employ potentially more powerful temporal features based on characteristics such as skin-color MHI, soft-weight assignment feature fusion [36], or optical flow. Some of the hyper-parameters used in the proposed approach are the focus of the experiments in Section 4. These hyper-parameters are optimized by a grid search to maximize the mean average precision over a validation set of the ICVL dataset. The temporal features in this paper are constructed using a $\tau_{max}$ value of 255, a $\tau_{min}$ value of 0, and a scalar threshold $\xi_{thr}$ of 30.

### 3.4 Multi-CNN Action Classifier

Three appearance-based temporal features are extracted from human action regions for BDI-CNN, MHI-CNN, and WAI-CNN, respectively. The first network, BDI-CNN, takes as input the BDI and captures the shape of the actor. The second network, MHI-CNN, operates on the MHI and captures the motion history of the actor. The third network, WAI-CNN, operates on the WAI and captures both the shape and the motion history of the actor. The three CNNs are trained sequentially for the task of action classification.

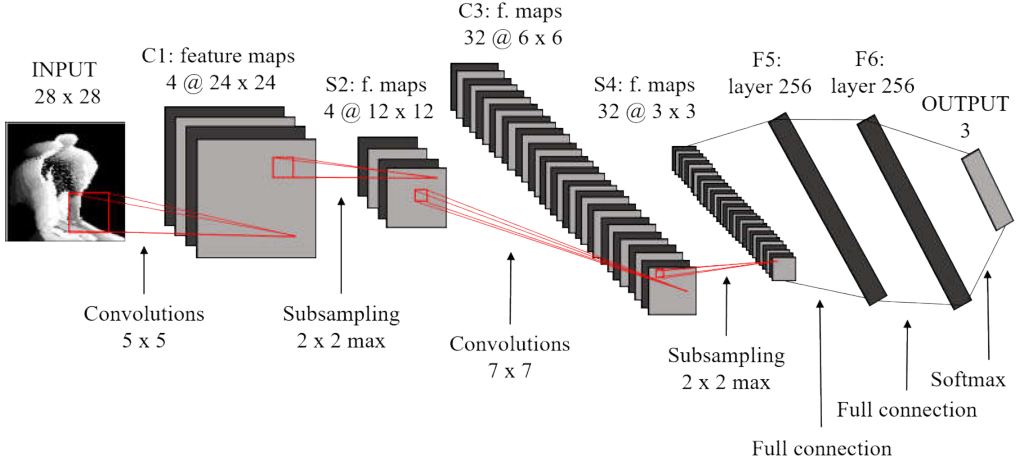

Figure 8: Architecture of a CNN. The architecture comprises two convolutional layers, two subsampling layers, two fully connected layers, and one softmax regression layer.

The selection of the optimal architecture of a CNN for a specific problem is challenging because this depends on the application. To keep the computation time low, we devised a light CNN architecture for real-time human action detection. The architecture of the CNN is shown in Figure 8 [19]. The architectures of BDI-CNN, MHI-CNN, and WAI-CNN are identical. Define $I(m)$ as an input image with size $m \times m$; $C(k, n, s)$ is a convolutional layer with a kernel size of $k \times k$, $n$ filters, a stride of $s$, and no padding. $Sm(k, s)$ is a subsampling layer of size $k \times k$ and stride $s$ using max pooling. *ReLUs* indicates rectified linear units, $FC(n)$ is a fully connected layer with $n$ neurons, and $D(r)$ is a dropout layer with a dropout ratio $r$. The architecture of the network is as follows: $I(28) - C(5,4,1) - ReLUs - Sm(2,2) - C(7,8,1) - ReLUs - Sm(2,2) - FC(256) - D(0.5) - FC(256) - D(0.5) - FC(|O_i|)$. The output layer consists of the same number of units as the number of sub-actions at the corresponding level of the descriptor. Finally, a softmax regression layer is added at the end of the network. If computational efficiency is not critical, one could use more complicated architectures [51], [52].

The weights in each layer are initialized using a zero-mean Gaussian distribution with a standard deviation of 0.01. The neuron biases in the convolutional layers and fully connected layers are initialized with the constant value of 1. Weight initialization is important in deep neural networks. Other works used transfer learning for initializing weights, especially in the absence of data. However, the advantage of transfer learning cannot be observed in this work owing the light architecture of the CNN and the abundant training data for each category on the ICVL dataset.

BDI-CNN, MHI-CNN, and WAI-CNN are trained using stochastic gradient descent (SGD) with a batch size of 256 examples, a learning rate of 0.001, a momentum of 0.9, and a weight decay of 0.0005, as in Krizhevsky et al. [51]. The networks are trained for 1 K iterations. The update rule of the weight is given by Eqs. 8 and 9

$$v_{i+1} := m \cdot v_i - \alpha \cdot \varepsilon \cdot w_i - \varepsilon \cdot \left\langle \frac{\partial L}{\partial w} \bigg| w_i \right\rangle_{D_i} \tag{8}$$

$$w_{i+1} := w_i + v_{i+1} \quad , \tag{9}$$

where $v$ is a momentum variable, $m$ is the momentum, $\alpha$ is the weight decay, $\varepsilon$ is the learning rate, $i$ is the iteration index, and $\left\langle \frac{\partial L}{\partial w} \big| w_i \right\rangle_{D_i}$ is the average over the $i$th batch $D_i$ of the derivative of the objective function with respect to $w$, as evaluated at $w_i$.

With a padding of 10 pixels in each dimension, the human action regions of BDI, MHI, and WAI are cropped to the bounding boxes and resized to 28×28. The average intensity values of BDI, MHI, and WAI are subtracted from the input temporal feature. Then, the features are flipped by mirroring the maps horizontally with a probability of 0.5.

### 3.5 Post-Processing

The discriminative action classifier composed of multiple CNNs, makes three predictions for each action, and the predictions are revised by post-processing to obtain the final decisions. As shown in Figure 1(c), the connection lines between two sub-actions at different levels of the descriptor indicate that the two sub-actions are independent of each other. No connection indicates an incompatible relation, in which the two sub-actions cannot occur together. For instance, the sub-actions *sitting* at the posture level and *walking* or *running* at the locomotion level cannot appear together. The multi-CNN classifier makes predictions at a single frame. The predictions of the classifier are checked by the connections of the sub-action descriptor. If the predictions conflict with the designed descriptor, the predictions are revised by Eqs. 10 and 11:

$$p_k = \arg\max_{\alpha_k} \left[ P(\alpha_k | \mathbf{x}_i, s_j) \right] \tag{10}$$

$$\arg\max_{\alpha_k} \left[ P(\alpha_k | \mathbf{x}_i, s_j) \right] = \arg\max_{\alpha_k} \left[ \frac{P(\mathbf{x}_i | \alpha_k, s_j) P(\alpha_k, s_j)}{P(\mathbf{x}_i, s_j)} \right]$$

$$\propto \arg\max_{\alpha_k} \left[ P(\alpha_k | s_j) P(s_j) \right] \quad , \tag{11}$$

where the new prediction, $p_k$ ($k \in \{1, 2, 3\}$), is given by sub-action $\alpha_k$ at the $k$th level of the descriptor that maximizes $P(\alpha_k|\mathbf{x}_i, s_j)$. $\mathbf{x}_i$ is the CNN feature vector of the person, and $s_j$ ($j \in \{1, 2, …, 12\}$) is one of the 12 camera scenes. To simplify the solution, we assume the distribution of $P(\mathbf{x}_i | \alpha_k, s_j)$ to be uniform and we calculate $P(\alpha_k, s_j)$ from the training data for each sub-action in the 12 camera scenes. It is noted that there is a relation between a sub-action and a scene; therefore, the joint probability of the action and the scene is not equal to the multiplication of the prior probability of sub-action $P(\alpha_k)$ and the prior probability of scene $P(s_j)$. In addition, the revised predictions are saved in the sub-action history memory on the basis of the human ID. Finally, temporal smoothing is used to reduce false predictions with regard to the saved sub-action history. This simple and efficient process works well, in practice, to reduce noise and decrease sensitivity.

## 4. Experimental Results

In this section, the proposed approach for real-time action detection in surveillance videos is evaluated in terms of the recognition rate and processing time. Systematic estimation of several hyper-parameters of the appearance-based temporal features is investigated. In addition, an ablation study of the appearance temporal features with the CNN-based approach is presented, and the results of the action detection are shown with the ICVL dataset, which is the only dataset suitable for multiple-individual action detection in surveillance videos. The average processing time was computed as the average time required to localize and recognize an actor in a test video. Meanwhile, experiments on the KTH dataset [5] were performed to compare the performance of the proposed method with those of other existing methods. The experimental results showed that appearance-based temporal features with a multi-CNN classifier effectively recognize actions in surveillance videos.

### 4.1 Evaluation Metrics

To quantify the results, we use the average precision at the frame-based *frame-AP* and at the video-based *video-AP*. The *frame-AP* was used in other approaches (e.g., object detection and image classification) at the frame-based evaluation, and *video-AP* provides an informative measurement for the task of action detection at the video-based evaluation.

- Frame-AP: Detection is correct if the intersection-over-union with the ground truth and detection area at that frame is greater than $\sigma$, and the action label is correctly predicted.
- Video-AP: Detection is correct if it satisfies the conditions of frame-AP on the spatial domain and the intersection-over-frames with the ground truth, and if the value for correctly predicted frames for one action is greater than $\tau$ in the temporal domain [47].

In addition, the mean average precision (mAP) for all action categories at the frame-based and the video-based measurement was used to evaluate the proposed approach because multiple actions can appear in one video, and the distribution of instances within each category is highly unbalanced on the validation and test set. An intersection-over-union threshold of $\sigma = 0.5$ and an intersection-over-frames threshold of $\tau = 0.5$ were leveraged in all methods and across the experiments. The hyper-parameters $n$ in Eq. 6 and $w_1$ and $w_2$ in Eq. 7 were defined through the following experiments.

### 4.2 Action Detection on ICVL Dataset

Challenging public datasets promote progress in the computer vision research area. There are many action public datasets which contain sports (*e.g.* UCF sports dataset) and user videos (*e.g.* THUMO'14 and ActivityNet). However, action categories in these datasets are totally different with the actions in real-world surveillance scenes. Moreover, the video in these datasets has one actor with one single action that labeled for the video, but in the surveillance video there are multiple actors with different actions at the same time. Therefore, some action recognition algorithms that focus on sports and user videos are not appropriate on the surveillance videos. This paper introduces the ICVL dataset to provide a better public benchmark and helps overcome the limitations of current action detection capabilities in real-world surveillance environments. In the following, the ICVL dataset and annotation information are presented.

Compared to existing surveillance datasets, the ICVL dataset has richer single-person actions, which were captured from real indoor and outdoor unscripted surveillance footage. One important characteristic of the ICVL dataset is that it includes different sub-actions of multiple individuals occurring simultaneously at multiple space locations in the same scene. There are diverse types of sub-action categories on the ICVL dataset: *sitting*, *standing*, *stationary*, *walking*, *running*, *nothing*, *texting*, *smoking*, and *others*. In terms of annotation, sub-action annotations are provided for each action, e.g., a person is marked simultaneously by three sub-actions: *standing*, *walking*, and *smoking*.

The ICVL dataset collected approximately 7 h of ground-based videos across 12 non-overlapping indoor and outdoor scenes, with a video resolution of 1280 × 640 at 15 Hz. Snapshots of 12 scenes are shown in Figure 9, including five indoor scenes and seven outdoor scenes. To avoid occlusion as much as possible, the authors of the dataset installed multiple models of high-definition video cameras on rooftops or high poles; therefore, the view angles of the cameras were toward dominant ground planes. For a new public dataset, the ICVL dataset provides a new benchmark for detecting the actions of multiple individuals. The ICVL dataset together with the annotations is publicly available.[1]

Annotating a large video dataset is a challenging work, especially while maintaining high quality and low cost. Annotations on the dataset include two types of ground truths: bounding boxes for persons, and sub-action labels for each action from each frame. Only the visible parts of persons are marked by whole and tight bounding boxes, and they are not extrapolated beyond occlusion by guessing. The annotation of the bounding boxes follows VATIC [53]. Automatic interpolation was used to recover the bounding boxes between key frames. Furthermore, the results were vetted and edited again manually. Once spatial bounding boxes were marked, temporal sub-actions were labeled for each action. One of the fundamental issues in sub-action labeling is deciding the starting moment and ending moment of an action. For instance, the action *standing up* has a posture conversion from *sitting* to *standing*. Deciding the confines between *sitting* and *standing* is ambiguous. To avoid ambiguity in action-labeling tasks, we labeled the actions on the basis of visual information and not of guesswork. For sub-action labeling, the approach studied by Yuen et al. [54] was referenced.

Table 1: Statistics of the ICVL dataset. The dataset focuses on action detection in video surveillance and was collected in real-world environments.

| Dataset | Resolution | Fps (Hz) | Duration (hours) | No. of subjects | No. of sub-action categories | No. of training videos | No. of validation videos | No. of test videos | No. of cameras | Camera type |
|---------|-----------|----------|------------------|-----------------|------------------------------|------------------------|--------------------------|---------------------|----------------|-------------|
| ICVL | 1280 × 640 | 15 | 7 | 1793 | 9 | 387 | 50 | 60 | 12 | Stationary ground |

The statistics of the dataset are summarized in Table 1. The ICVL dataset consists of 387 training videos, 50 validation videos, and 60 test videos (5 videos for each camera in validation and test set). To increase the difficulty of the dataset, the authors of the dataset used only camera 01 to camera 10 for the training and validation set and included unseen contexts (camera 11 and camera 12) for the test set. In the following experiments, this study focused on the recognition of eight sub-action categories (*sitting*, *standing*, *stationary*, *walking*, *running*, *nothing*, *texting*, and *smoking*). Each sub-action was classified in a one-against-the-rest manner, and a number of negative samples (*others*) were selected from the actions that were not in these eight categories. Human locations on the training set were manually labeled with bounding boxes, whereas a human detector was used to automatically localize each person on the test set. For efficiency, the videos of the ICVL dataset were resized to 640 × 320. The evaluation used the ICVL test set, and all hyper-parameters were analyzed on the validation set.

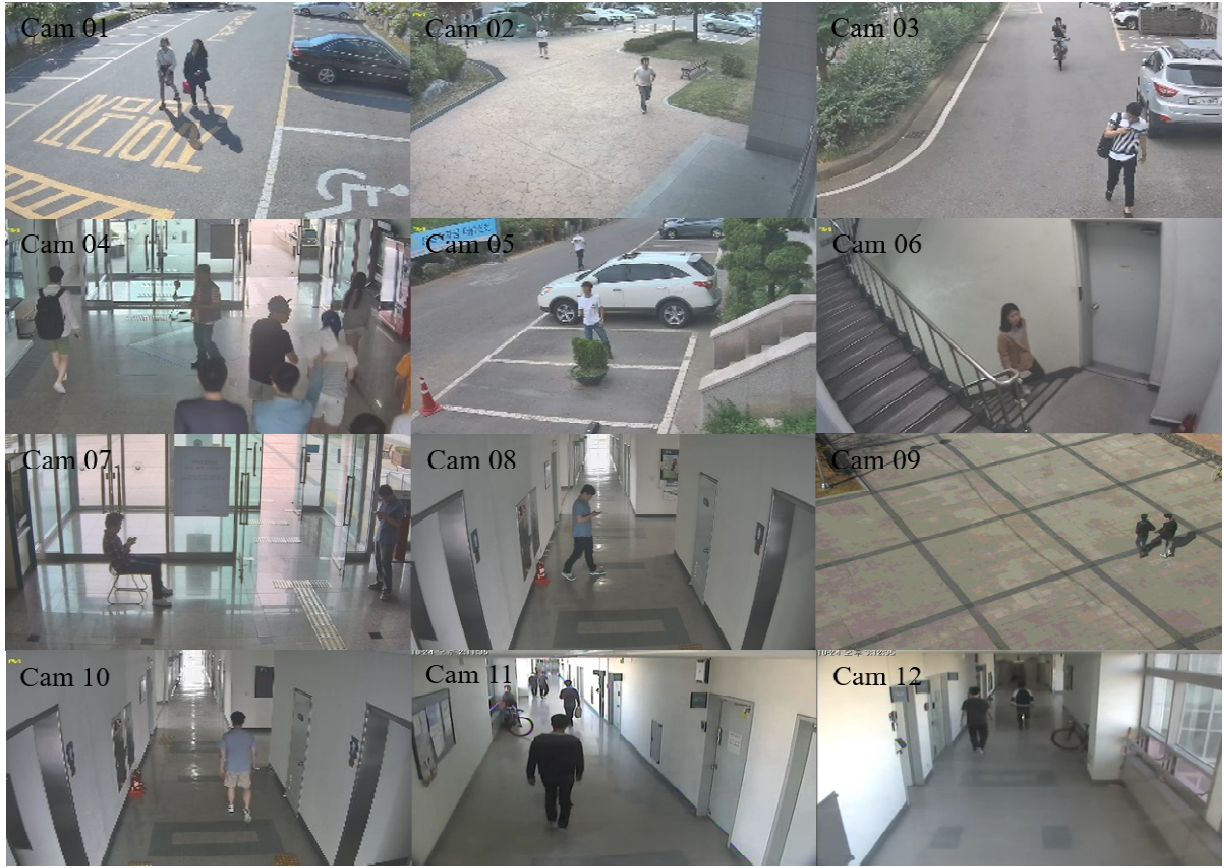

Figure 9: Example scenes from 12 different cameras on the ICVL dataset. Seven example scenes are outdoor scenes, and another five examples are indoor scenes. Camera 08 and Camera 10 were in a similar context, where two cameras were set on different floors of the same building. The videos recorded by Camera 09 were excluded from the evaluation of the proposed method because only a few persons appeared in this scene.

MHI-CNN operates on MHI to capture the motion history of an actor. MHI encodes sequential frames as memory capacity to represent actions. However, deciding the number of frames *n* in Eq. 6 is a highly action-dependent issue. MHI with a large *n* covers a wide range of action history, but it is insensitive to current actions. However, MHI with a small *n* puts the focus on the recent actions and ignores past actions. In this paper, the number of frames in the MHI was defined by performing a grid search from 5 to 50 frames with an interval of 5. Figure 10 plots the classification accuracy (mAP) at the frame-based and the video-based measurement for the sub-actions at the locomotion level of the sub-action descriptor. Peaks and a plateau around the peaks of *n* were observed at both the frame-based and the video-based measurement. With *n* equal to 25 frames, MHI-CNN was able to achieve a consistent performance boost from 1% to 2% of the mAP at the frame-based and video-based measurement. It is worth emphasizing that recognizing unscripted actions is such a challenging task that a 2% absolute performance improvement is quite significant. This evidence indicates that correctly recognizing one action would need approximately 2 s (15 fps in the ICVL videos). This result was also efficient in WAI-CNN for the third level of the sub-action descriptor. For the remainder of the experimental results, $n = 25$ was used in MHI and WAI.

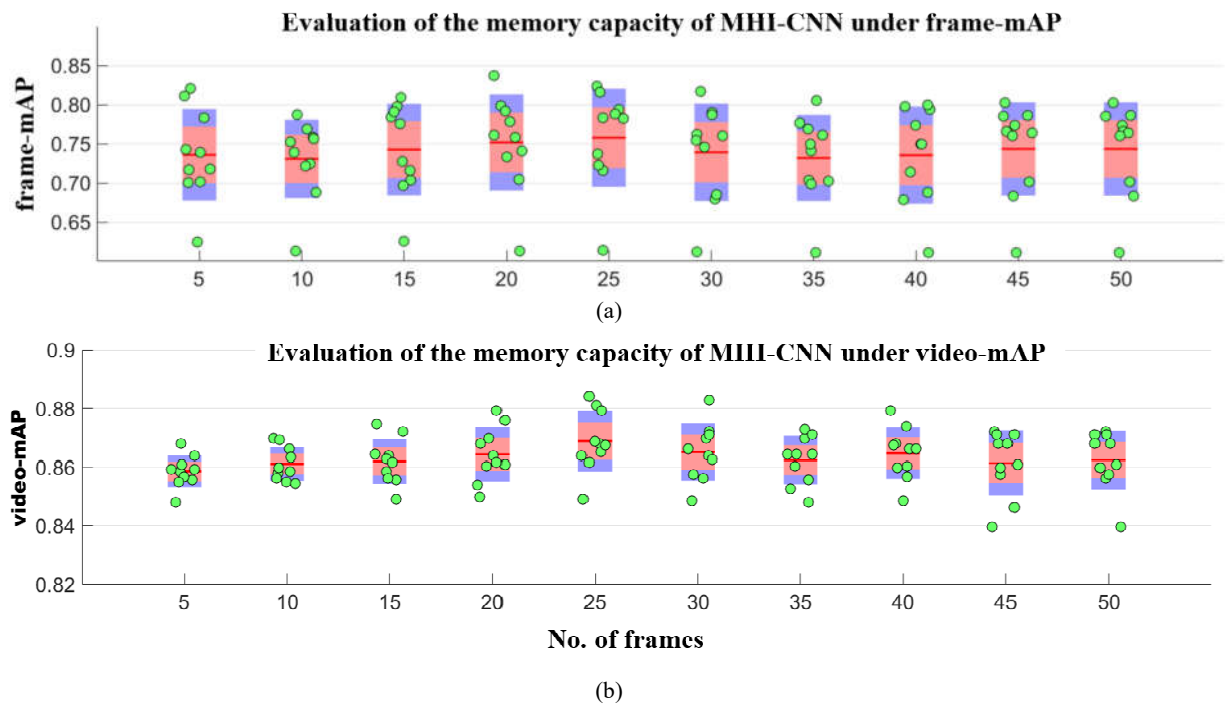

(a)

(b)

Figure 10: Memory capacity in MHI for the locomotion level of the sub-action descriptor. (a) is the results of the mAP at the frame-based measurement, and (b) is the results of the mAP at the video-based measurement. The green circles are drawn while training MHI-CNN from 100 to 1K iterations with an interval of 100. The circles lie over a 1.96 standard error of the mean in red and 1 standard deviation in blue. The baseline accuracy at $n = 10$ is given by encoding the temporal features.

The third network, WAI-CNN, operates on the linear combination of the static and motion history cues and captures specific patterns of the actor. As in Eq. 7, WAI is the weighted average of BDI and MHI. An ablation study of the proposed approach at the gesture level is presented by evaluating the performance of the two appearance-based temporal features, BDI and MHI, and their combination. Table 2 shows the results of each temporal feature with CNN. The best results are highlighted in bold. MHI-CNN always performed a bit better than BDI-CNN at both frame-AP (45.5% vs. 38.7%, respectively) and video-AP (55.3% vs. 51.2%, respectively). This is due to the fact that the motion history cues are very informative for some action patterns. As shown in the ablation study, it is apparent that WAI-CNN, which is a linear combination of BDI and MHI, performed significantly better for all sub-actions. WAI-CNN was even better than MHI-CNN at both frame-AP (59.7% vs. 45.5%, respectively) and video-AP (65.2% vs. 55.3%, respectively). It is clear that shape and motion history cues are complementary toward action recognition. $n = 25$, $w_1 = 0.5$, and $w_2 = 0.5$ were used in this experiment.

Table 2: Results of the ablation study on the gesture level of ICVL dataset. Frame-AP and video-AP are reported for BDI-CNN, MHI-CNN, and WAI-CNN. WAI-CNN performed significantly better under both metrics, showing the significance of the combined cues for the task of gesture-level sub-action recognition. The leading scores of each label are displayed in bold font.

| frame-AP (%) | nothing | texting | smoking | mAP |
|---|---|---|---|---|
| BDI-CNN | 58.7 | 47.1 | 10.3 | 38.7 |
| MHI-CNN | 64.6 | 58.2 | 13.7 | 45.5 |
| WAI-CNN | **81.6** | **70.6** | **26.9** | **59.7** |
| video-AP (%) | | | | |
| BDI-CNN | 77.2 | 54.7 | 21.7 | 51.2 |
| MHI-CNN | 70.2 | 63.8 | 31.9 | 55.3 |
| WAI-CNN | **82.1** | **75.3** | **38.2** | **65.2** |

The performance of WAI-CNN with respect to the weights in WAI (Eq. 7) was further evaluated. Figure 11 shows the mAP across sub-actions at the gesture level of the sub-action descriptor at the frame-based and video-based measurement with regard to varying weights on WAI and training iterations of the action-CNN. It is noted that using $w_1 = 1.0$ and $w_2 = 0.0$ is equivalent to using a shape cue only, and using $w_1 = 0.0$ and $w_2 = 1.0$ is equivalent to using only a motion history cue. In addition, we can see that a significant improvement can be achieved as the shape and motion cues are used together, as shown in Table 2. In Figure 11, it is noted that $w_1 = 0.6$ and $w_2 = 0.4$ outperformed the other weight combinations under both metrics. In particular, $w_1 = 0.6$ and $w_2 = 0.4$ show a significant improvement of (relatively) 7%, on average, at the frame-based measurement and (relatively) 11%, on average, at the video-based measurement beyond $w_1 = 0.5$ and $w_2 = 0.5$. This implies that the shape cue is more important than the motion history cue in WAI and is quite different from the results in Table 2. One possible explanation for this finding is that the motion history cue is more informative than the shape cue if they are used individually. However, after combining them, the shape cue contributes much more than the motion history cue for gesture-level action recognition. For the remainder of the experimental results, $w_1 = 0.6$ and $w_2 = 0.4$ were used in WAI.

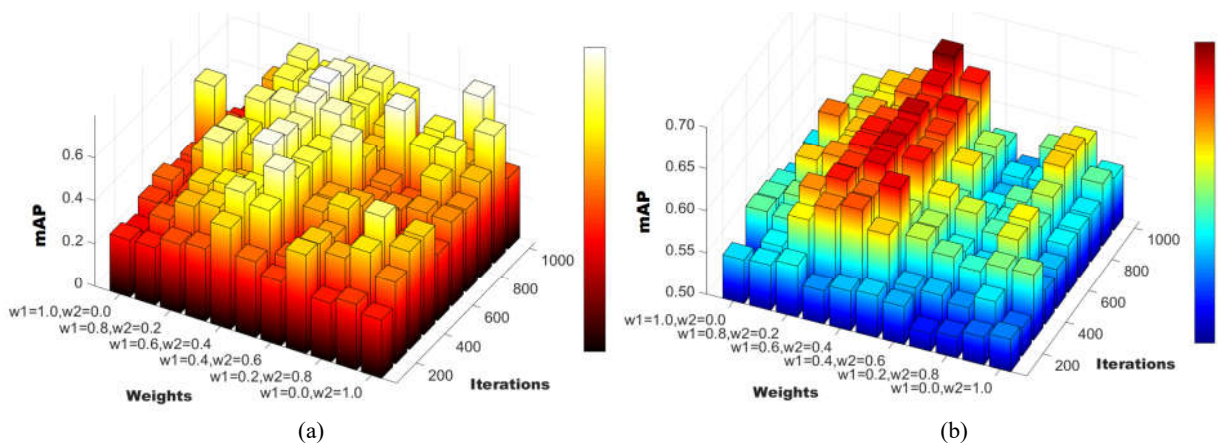

Figure 11: Recognition results with regard to varying weights of WAI and training iterations on WAI-CNN. (a) mAP of WAI-CNN at the frame-based measurement, and (b) mAP of WAI-CNN at the video-based measurement

To evaluate the effectiveness of the action-detection model, we included the full confusion matrixes as a source of additional insight. Figure 12 shows that the proposed approach achieved a mAP of 76.6% at the frame-based and 83.5% at the video-based measurement. The proposed method was able to get most of the sub-action categories correct, except for *smoking*. The poor performance with *smoking* was due to the fact that WAI-CNN captured a specific action cue, i.e., the hand-to-mouth pattern (Figure 7); however, that pattern does not appear when the person keeps the hand on the mouth and inhales. In other words, WAI-CNN for *smoking* without the hand-to-mouth pattern was almost always considered as *nothing*. Addressing this problem by using object localization and classification is a very challenging task. However, most of the misclassified samples are hard to recognize, even by a human. The results of the experiment show that a sub-action descriptor can eliminate many misclassifications by dividing one action into many sub-actions that are not at the same levels.

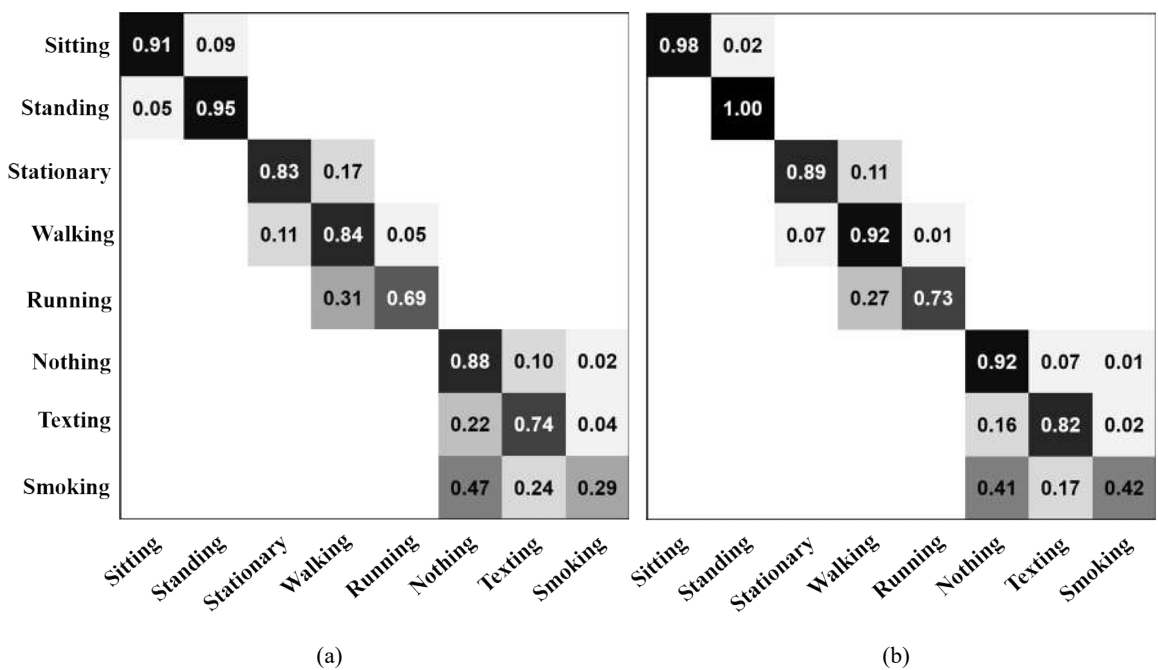

(a)          (b)

Figure 12: Confusion matrixes of the ICVL dataset at the frame-based and video-based measurement for the action-detection task when using appearance-based temporal features with a multi-CNN classifier. The horizontal rows are the ground truth, and the vertical columns are the predictions. Each row was normalized to a sum of 1. (a) The confusion matrix at the frame-based measurement, and (b) the confusion matrix at the video-based measurement

Inspired by the capability of a deep neural network, one may want to composite multiple sub-actions to solve the representation problem of an action. Sadeghi and Farhadi et al. [55] and Xu et al. [56] introduced an actor-action tuple as a visual phrase by considering multiple different categories of actors undergoing multiple different categories of actions. For example, person riding, dog lying, and car jumping, etc. They believe that detecting visual phrases are much easier than independently detecting participated objects because the appearance of objects may change when they participate in relations.

To assess the effectiveness of a multi-CNN classifier with a sub-action descriptor, we constructed 10 visual phrases to train CNN directly and compare them with the proposed method. According to Figure 1(c), these visual phrases are *sitting with nothing*, *standing with nothing*, *walking with nothing*, *running with nothing*, *sitting with texting*, *standing with texting*, *walking with texting*, *sitting with smoking*, *standing with smoking*, and *walking with smoking*. For simplicity, *sitting while stationary* is shortened to *sitting*, *standing while stationary* is shortened to *standing*, and *standing with walking* is shortened to *walking*. *Running with texting* and *running with smoking* were excluded because they are not available in the ICVL dataset. Different appearance-based temporal features, BDI, MHI, and WAI, can be applied as an input of CNN. We call them visual phrase BDI-CNN, MHI-CNN, and WAI-CNN, respectively. Table 3 shows the results of the comparison of the performance of the methods with visual phrases on the ICVL dataset. Notably, the proposed multi-CNN classifier with a sub-action descriptor outperformed the visual phrase methods. A possible explanation of the above result is that Sadeghi and Farhadi [55] and Xu et al. [56] considered multiple actor-action interactions, whereas we focued on the context of human actors and their sub-actions. Experiments of conditional model in [56] indicate similar conclusion that knowing actor categories can help with action inference. In addition, the number of human-action visual phrases grows exponentially, but the very large number of phrases that can be built has very few sub-actions. In this way, it decreases the number of sub-actions and reduces misclassifications.

Table 3: Results of the comparison of frame-mAP performance using methods with visual phrases on the ICVL dataset. The leading score of mAP is displaced in bold font.

| Method | Frame-mAP (%) |
| --- | --- |
| Visual phrase BDI-CNN | 42.7 |
| Visual phrase MHI-CNN | 51.1 |
| Visual phrase WAI-CNN | 56.6 |
| Multi-CNN classifier with a sub-action descriptor | **76.6** |

Figure 13 and Figure 14 gives some qualitative localization and recognition results using the proposed approach on the test set of the ICVL dataset. Each block corresponds to a video from a different camera. A red bounding box indicates the action in the region of interest and predicted sub-action labels with corresponding icons; human IDs are in the top-left corner of the bounding boxes, and random color trajectories are shown in the video. The second image of the third example is an incorrect prediction, with the true gesture label being *smoking*.

To provide an evaluation of the processing time, we established an experimental environment on a computer with an Intel Core i7-3770 CPU at 3.40 GHz with two 4 GB RAM modules. The average processing time was computed as the average time required to localize actions in the regions of interest and to recognize actions in a test set. The input video was resized to 640 × 320, and the processing time was tested on 33 videos shown in Table 4.

The average processing time for one frame was 41.83 ms. Tracking by motion detection was the most time-consuming part. The three appearance-based temporal features were very fast and are suitable for real-time surveillance applications. A light CNN architecture took just 3.66 ms to predict three sub-actions, and the other processes (e.g., initialization, displaying the results, etc.) took 13.06 ms. As can be seen above, the proposed action detection model ran at around 25 fps.

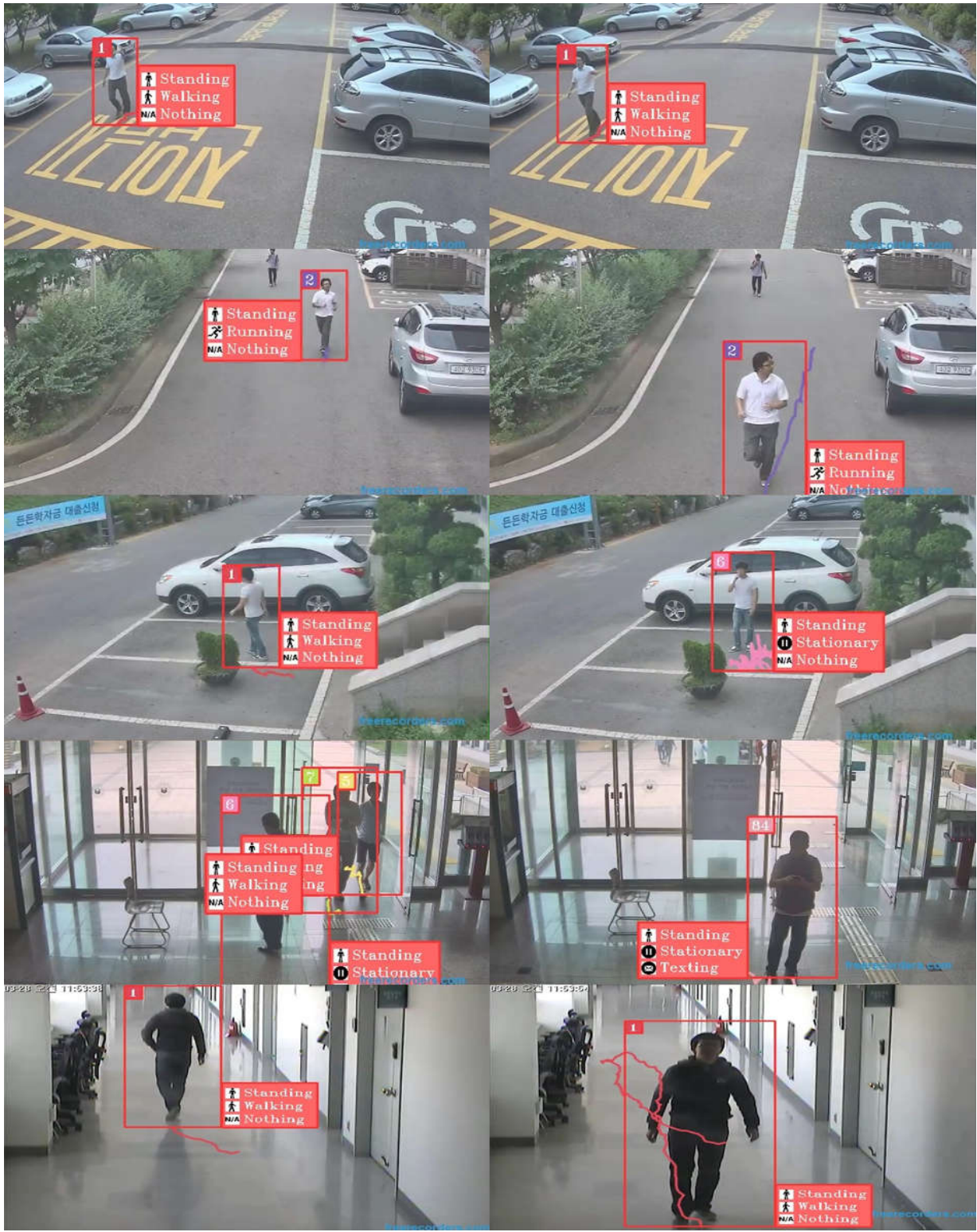

Figure 13: Examples of action localization and recognition results from the ICVL dataset. Each block corresponds to a video from a different camera. Two frames are shown from each video. The second frame of the third example is an incorrect prediction, with the true gesture label being *smoking*.

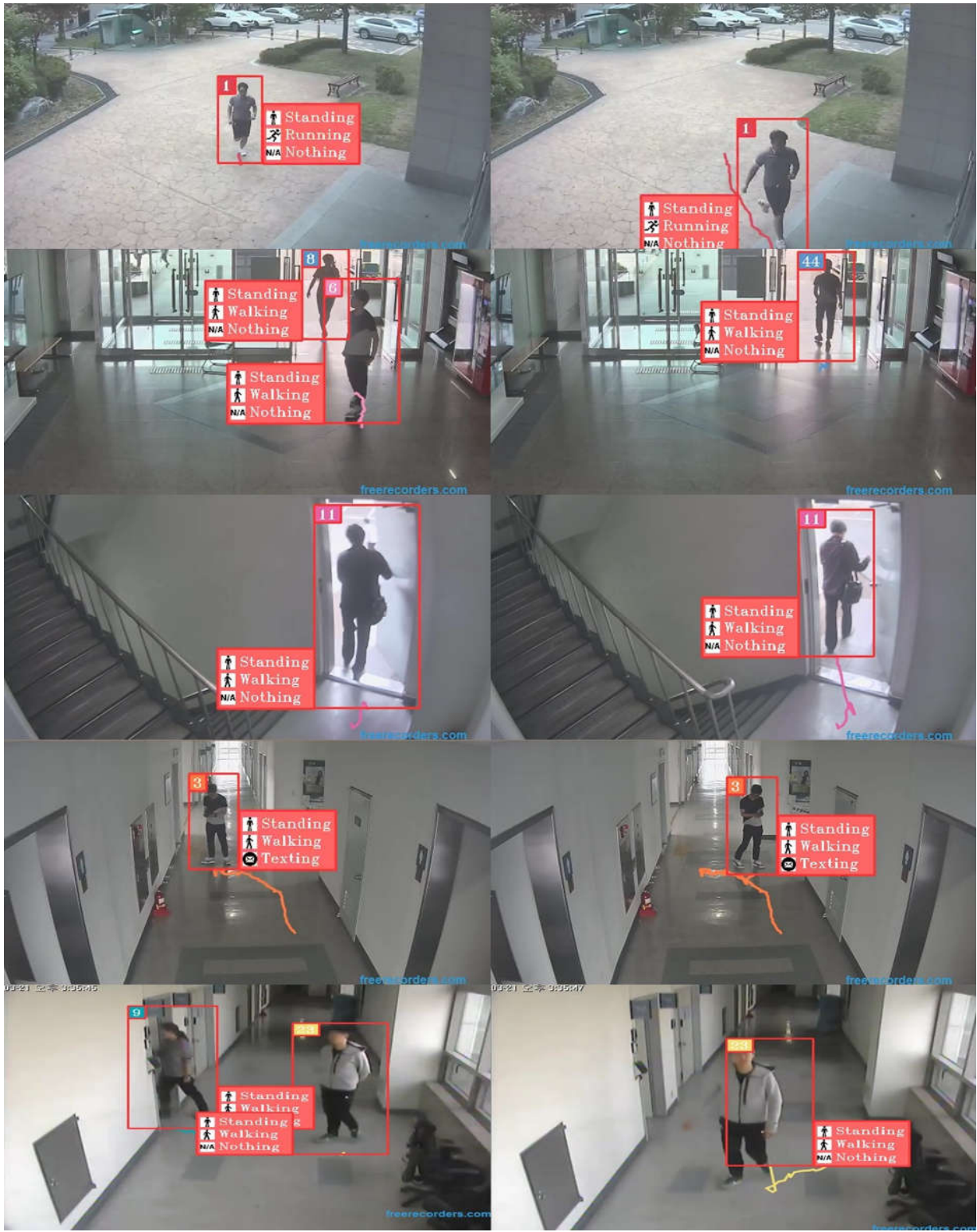

Figure 14: Examples of action localization and recognition results from the ICVL dataset. Each block corresponds to a video from a different camera. Two frames are shown from each video.

Table 4: Average processing time of the proposed action detection model

| Module | Motion detection | Detector | Tracker | BDI | MHI | WAI | CNNs | Post-processing | Others | Overall |
|---|---|---|---|---|---|---|---|---|---|---|
| Processing time (ms) | 11.30 | 12.60 | 0.23 | 0.23 | 0.83 | 0.10 | 3.66 | 0.03 | 13.06 | 41.83 |

### 4.3 Action Recognition on the KTH Dataset

The proposed method was evaluated on the KTH dataset [5], which consists of six action categories, namely, *boxing*, *hand-clapping*, *hand-waving*, *jogging*, *running*, and *walking*, performed by 25 subjects. Similar to Figure 1(c), the sub-action descriptor of KTH can be seen in Figure 15. For instance, the action *boxing* consists of sub-action *stationary* at the locomotion level and sub-action *boxing* at the gesture level, and the action *running* consists of sub-action *running* at the locomotion level and sub-action *nothing* at the gesture level. There is only one sub-action, *standing*, at the posture level. We just neglected sub-action *standing* here.

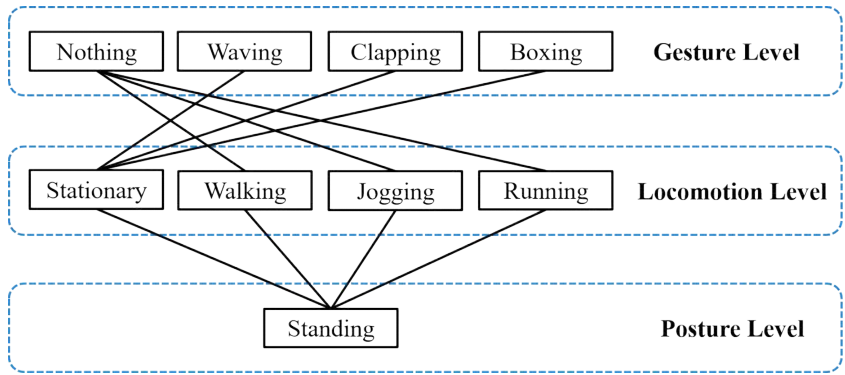

Figure 15: Structure of the sub-action descriptor on the KTH dataset.

The frame rate of KTH is 25 fps and the resolution is 160 × 120. The ground truth did not supply the bounding box information of the human area, and, therefore, the multi-CNN classifier was trained using a full image as an input. In the test phase, the proposed method did not run the human detection and tracking modules. As in [5], we used the data for eight subjects (person11, 12, 13, 14, 15, 16, 17, 18) for the training, the data for eight subjects (person01, 04, 19, 20, 21, 23, 24, 25) for the validation, and the data for the remaining nine subjects (person02, 03, 05, 06, 07, 08, 09, 10, 22) for the testing. The hyper-parameters were optimized on the validation set. The majority voting was used to produce the labels for a video sequence based on the prediction for individual frames. A video was correct if and only if the sub-actions at the locomotion and gesture level were correct.

A performance comparison between the proposed method and the state-of-the-art results on the KTH is reported in Table 5. Some results, which used leave-one-out cross-validation (24 subjects for the training and the 1 subject for the test) are not reported here. The performance obtained by the proposed multi-CNN classifier with a sub-action descriptor was among the top on the KTH dataset. Moreover, the proposed method ran at around 86 fps on the KTH dataset; however, other existing works did not consider the processing time, which is critical in surveillance applications. Figure 16 shows the confusion matrix of the proposed method on the KTH dataset.

Table 5: Comparison of the performance on the KTH dataset. The leading score and fps are displayed in bold fond.

| Method | mAP (%) | Processing Time (fps) |
| --- | --- | --- |
| Kovashka and Grauman [57] | 94.5 | – |
| Wang et al. [9] | 94.2 | – |
| Gilbert et al. [58] | 94.5 | – |
| Baccouche et al. [59] | 94.4 | – |
| Zhang et al. [60] | 94.1 | – |
| Kaâniche and Brémond [61] | 94.7 | – |
| Ji et al. [44] | 90.2 | – |
| Bilinski et al. [62] | 94.9 | – |
| Chen et al. [63] | 94.4 | – |
| Selmi et al. [64] | 95.8 | – |
| Liu et al. [65] | 95.0 | – |
| Multi-CNN classifier with a sub-action descriptor | **96.3** | **86** |

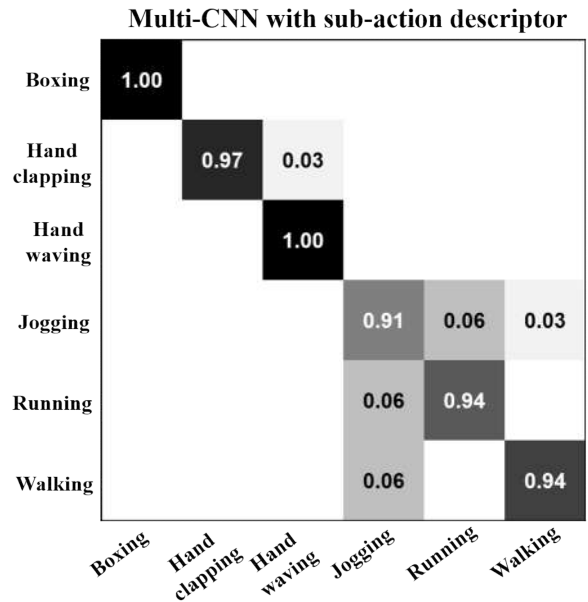

Figure 16: Confusion matrix of the multi-CNN classifier with a sub-action descriptor on the KTH dataset. The horizontal rows are the ground truth, and the vertical columns are the predictions. Each row was normalized to a sum of 1.

Direct training with different appearance-based temporal features for six categories was also evaluated on the KTH. Table 6 shows the per-class and mAP performance on the KTH. The proposed method outperformed the others in all of the categories. Owing to the ambiguousness of MHI for interior motions, the results for sub-action *jogging*, *running*, and *walking* should be improved further.

Table 6: Per-class breakdown and mAP on the KTH dataset. The leading scores of each label are displayed in bold font.

| Method | Boxing | Hand-clapping | Hand-waving | Jogging | Running | Walking | mAP (%) |
|---|---|---|---|---|---|---|---|
| BDI-CNN | 86.11 | 80.56 | 72.22 | 72.22 | 80.56 | 75.00 | 77.78 |
| MHI-CNN | 88.89 | 91.67 | 83.33 | 86.11 | 88.89 | 86.11 | 87.50 |
| WAI-CNN | 94.44 | 94.44 | 91.67 | 86/11 | 91/67 | 88.89 | 91.20 |
| Proposed method | **100** | **97.22** | **100** | **91.67** | **94.44** | **94.44** | **96.30** |

## 4.4 Discussions

The following important conclusions can be drawn from the above experimental results. Shape cues (BDI) lead to captured silhouette features from the spatial domain and can effectively identify the posture of an actor. They can provide over 95% mAP at the posture level of the sub-action descriptor. Motion history cues (MHI), even simple and fast temporal features, are of crucial importance for recognizing sub-actions of the locomotion level: *stationary*, *walking*, and *running*. However, deciding the memory capacity of MHI is a highly action-dependent issue. As determined from a large set of experiments, correctly recognizing one action takes approximately 2 s with the ICVL dataset. The combination of shape and motion history cues (WAI), when the weighted average was used with them, can provide further improvement in performance for the gesture level of the sub-action descriptor. Shape and motion history cues are complementary for gesture-level sub-action recognition. From an ablation study, it was noted that the motion history cue was more informative than the shape cue if the two were used individually. However, after combining them, the shape cue contributed much more than the motion cue for gesture-level sub-action recognition. The actor-action visual phrase provides a significant gain over object detectors coupled with existing methods modeling human-object interactions [55], [56] because the appearance of the objects may change when they participate in relations. However, for human activity, the sub-actions at different levels can happen independently with less appearance variations in the proposed sub-action descriptor. Moreover, it solves the problem of how much the visual phrases grow exponentially in the number of objects.

## 5. Conclusions

This paper presented a novel real-world surveillance video dataset and a new approach to real-time action detection in video surveillance system. The ICVL dataset will stimulate research on multiple action detection in the years ahead. Extensive experiments demonstrated that a sub-action descriptor delivers a complete set of information about human actions and significantly eliminates misclassifications by the large number of actions that are built by few independent sub-actions at different levels. An ablation study was presented about on the basis of the ICVL dataset and showed the effect of each temporal feature when considered separately. Shape and motion history cues are complementary, and using both leads to a significant improvement in action recognition performance. In addition, the proposed action detection model simultaneously localizes and recognizes the actions of multiple individuals at low computational cost with acceptable accuracy. The model achieved state-of-the-art result on the KTH dataset and ran at around 25 fps on the ICVL dataset and at 86 fps on the KTH dataset, which is suitable for real-time surveillance applications. However, the performance for recognizing gesture-level sub-action (e.g., *smoking*) was not adequate. A promising direction for future work is to extend our framework to learn the joint space of the sub-action descriptor. In addition, more powerful temporal feature methods, such as a skin-color MHI or optical flow, and other deep architectures of CNNs are being considered.

## Acknowledgment

This work was supported by a grant from the Institute for Information and Communications Technology Promotion (IITP) funded by the Korean government (MSIP) (B0101-15-1282-00010002, Suspicious pedestrian tracking using multiple fixed cameras).